# Attention-stacked Generative Adversarial Network (AS-GAN)-empowered Sensor Data Augmentation for Online Monitoring of Manufacturing System

Yuxuan Li, Chenang Liu*

The School of Industrial Engineering & Management, Oklahoma State University, Stillwater, OK

*Corresponding author. chenang.liu@okstate.edu

**Abstract:** Machine learning (ML) has been extensively adopted for the online sensing-based monitoring in advanced manufacturing systems. However, the sensor data collected under abnormal states are usually insufficient, leading to significant data imbalanced issue for supervised machine learning. A common solution is to incorporate data augmentation techniques, i.e., augmenting the available abnormal states data (i.e., minority samples) via synthetic generation. To generate the high-quality minority samples, it is vital to learn the underlying distribution of the abnormal states data. In recent years, the generative adversarial network (GAN)-based approaches become popular to learn data distribution as well as perform data augmentation. However, in practice, the quality of generated samples from GAN-based data augmentation may vary drastically. In addition, the sensor signals are collected sequentially by time from the manufacturing systems, which means sequential information is also very important in data augmentation. To address these limitations, inspired by the multi-head attention mechanism, this paper proposed an attention-stacked GAN (AS-GAN) architecture for sensor data augmentation of online monitoring in manufacturing system. It incorporates a new attention-stacked framework to strengthen the generator in GAN with the capability of capturing sequential information, and thereby the developed attention-stacked framework greatly helps to improve the quality of the generated sensor signals. Afterwards, the generated high-quality sensor signals for abnormal states could be applied to train classifiers more accurately, further improving the online monitoring performance of manufacturing systems. The case study conducted in additive manufacturing also successfully validated the effectiveness of the proposed AS-GAN.

# 1. Introduction

## 1.1 Background and motivation

In advanced manufacturing systems, sensor-based process monitoring techniques become more and more popular to understand the real-time process quality and potentially reduce the costs of system operation [1]. Specifically, in recent years, machine learning (ML)-based approaches are widely applied to enable effective online monitoring due to their powerful functionality and capability [2, 3]. These ML-based approaches are usually trained offline by the collected data, i.e., the multi-channel sensor signals collected during the operations, and then the trained models will be applied for online monitoring. However, in practice, most of the sensor signals are usually collected when the machines are under the normal state. Under such circumstances, the sensor signals collected under the abnormal state may not be sufficient for training ML models, especially for the supervised ML models, which is called data imbalanced issue [1-4]. It may lead to the significant monitoring bias when applying these trained ML models online. To address the data imbalanced issue in the manufacturing systems, according to the literature, data augmentation is a common direction to mitigate this issue [5-7].

To augment the abnormal state sensor signals effectively, one of the most critical aspects is to well learn the distribution of the sensor signals. Due to the complex structures of the sensor data, highly nonlinear underlying patterns may exist in the sensor signals. In addition, the sequential information is also necessary to be considered in learning the underlying distribution, since the sensor signals are usually collected sequentially by time [8, 9]. Therefore, the performance of the conventional data augmentation techniques, such as adding noise and the resampling techniques, may be not satisfactory [7]. Recently, with the fast-development of machine learning, the neural network-enabled techniques, such as the family of generative adversarial network (GAN) [10], are also leveraged to advance the sensor data augmentation [6]. GAN-based approaches utilize generators to augment the data and apply discriminators to distinguish the generated artificial and actual real data. When the training of GAN converges, the generator should be able to generate effective artificial data following the actual data distribution. Due to its powerful functionality to obtain a large number of synthetic samples (i.e., the generated artificial data) from the learnt distribution, many variants, such as the multivariate anomaly detection with GAN (MAD-GAN) [11] and augmented time-regularized GAN (ATR-GAN) [6], have been proposed to for advancing manufacturing systems, which also inspire this work. However, these GAN-based sensor data augmentation approaches still have critical shortcomings. For instance, MAD-GAN [11] considers the unsupervised cases without considering the existing label information. Besides, ATR-GAN [6] captures the sequential information indirectly by calculating temporal distance, and thus the quality of generated samples from GAN-based framework may still vary by cases. Therefore, how to better ensure the quality of the generated sensor data from the GAN model remains challenging.

## 1.2  Objective and contribution

Specifically, to augment the samples following sequential order, a natural idea is to incorporate a unique neural network architecture which could accomplish this task (i.e., sufficiently consider the sequential patterns of the sensor signals) in the GAN-based framework. Under such circumstances, the emerging neural network architecture, namely, the transformer [12], could be an appropriate option due to its powerful functionality for sequential-based generation. Compared with some other neural networks such as long-short-term memory (LSTM) [13], transformers could better consider long-range dependencies without the vanishing gradient problem. Hence, transformer has become one of the most popular tools in natural language processing (NLP), such as language translation [14] and named entity recognition [15]. Specifically, it successfully enables the effective generation of sequential values by incorporating the multi-head attention mechanism and the encoder-decoder architecture [16]. Notably, the multi-head attention mechanism is the key component in transformer to capture the sequential information. In this framework, each head aggregates the input sequence to measure the relationship between each point in the data. Then such multiple heads will be summarized together to estimate the attention scores, which are the weights to show the relevant information between different points of the input sequence.

Due to its strong capability on understanding the sequential data, transformer has also been incorporated to the GAN-based learning frameworks for time series data [17-19]. However, the existing GAN-based works with transformer for sensor data analytics in manufacturing systems are still very limited [18]. Besides, the transformer usually has a complicated structure and requires plenty of data for training, which is hard to be satisfied in manufacturing systems, as the abnormal states are usually rare [6]. Therefore, as shown in Figure 1, to be more convenient and efficient, the multi-head attention mechanism in transformer is extracted and integrated in this study, instead of directly applying the entire transformer structure. Consequently, less amount of training data is required to capture the sequential information. Since the output of multi-head attention mechanism is the attention scores, a regular multilayer perceptron (MLP)-based generator is also applied to generate samples with the help of attention scores (Figure 1).

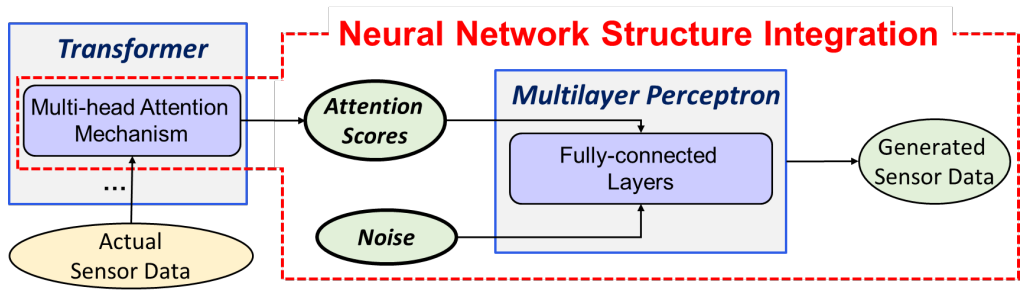

*Figure 1: A demonstration of the designed neural network structure integration.*

With the above-mentioned idea, this paper proposes a new GAN-based data augmentation approach termed attention-stacked generative adversarial network (AS-GAN). It stacks the attention scores obtained from different heads as well as stacks fully-connected layers to generate samples. In this proposed AS-GAN, a novel generator incorporating attention-stacked framework is established, where the specific contributions of the proposed AS-GAN consist of two main aspects: (1) a multi-head attention mechanism is integrated in the proposed attention-stacked generator to learn the nonlinear patterns in sensor signals and obtain the overall attention score for each sequential sample; and (2) a framework to stack multi-head attention mechanism and multilayer perceptron is further designed to generate high-quality sensor signals with the help of overall attention score.

The rest of this paper is structured as follows. The recent literatures related to data augmentation approaches are discussed in Sec. 2, followed by the proposed research methodology (i.e., AS-GAN) in Sec. 3. Then a simulation study and a real-world case study are applied to validate the effectiveness of the proposed method in Sec. 4. At last, conclusions and future works are drawn in Sec. 5.

## 2. Literature Review

The studies related to manufacturing system monitoring are briefly introduced first in Sec. 2.1. Afterwards, the existing data augmentation techniques for sensor data are summarized and the research gaps are also identified in Sec. 2.2.

### 2.1 Manufacturing system online monitoring

In recent years, the monitoring methods of manufacturing system are gradually developed to reduce the cost and improve the sustainability of manufacturing industries [17]. The monitoring techniques could be summarized into two categories: direct monitoring methods and indirect monitoring methods [21]. The direct measurements require to visualize the manufacturing systems for monitoring, such as illumination and cutting fluid. However, such measurements may interfere with the monitoring system and the machine tools, which may cause stability issues [21]. Hence, the monitoring systems utilizing indirect measurements, such as force and acceleration, developed rapidly in recent years [22]. Since sensors are the physical devices which could obtain electrical output by responding to the physical stimulus, they are widely applied to acquire the data as indirect measurements [5]. Nowadays, the sensors are widely utilized in many different manufacturing systems, such as additive manufacturing [9], laser-welding system [23], and milling system [24]. For instance, accelerometers are applied in the additive manufacturing to collect the vibration signals [9]. The collected vibration signals could be analyzed to monitor the state of manufacturing systems since the sensor signal changes could reflect the transition between normal and abnormal states of machine. Following this direction, extensive studies regarding the sensor data-driven online monitoring approach development have been conducted. For instance, Tao *et al.* [25] discussed how the data-driven strategies support smart manufacturing with the incorporation of big data. Besides, Shi *et al.* [26] summarized the innovative methodologies of in-process quality improvement and discussed the future direction in leveraging machine learning tools. Specifically, the manufacturing systems are usually mainly in control, which means the data collected when the manufacturing systems are out of control is limited. Then due to the sparsity of influenced sensor signals, monitoring all the sensor profiles may cause significant bias [27]. Therefore, data augmentation approaches, which could increase the number of insufficient influenced sensor signals, can potentially play an important role in sensor-based online monitoring in manufacturing.

**2.2 Data augmentation approaches for sensor data**

The most common expressions of sensor data are sequential data. Hence, the data augmentation approaches for sequential data are discussed, which are comprehensively reviewed in [28]. Specifically, the existing data augmentation approaches for sequential data could be summarized into two categories: statistics-based data augmentation approaches [29], and machine learning-based data augmentation approaches [30]. In statistical data augmentation approaches, synthetic minority oversampling technique (SMOTE) is a popular approach due to its convenience and intuitiveness [31]. This approach obtains new artificial samples by averaging the distance between different samples. In addition, a lot of extensions, such as Borderline-SMOTE (B-SMOTE) [32], have been proposed to further improve the rules for the selection of instances to be oversampled. However, the weighted average of existing samples generated from SMOTE-based approaches could not consider the nonlinear and temporal patterns among the data. Specifically, to consider the temporal patterns in the data, a popular approach is the dynamic time warping barycenter averaging (DBA) [33]. It quantifies the similarity between time series data using dynamic time warping (DTW) [34]. However, the computational cost of DBA is extremely high, which is not suitable for the online monitoring application. In addition, decomposition-based approaches are widely applied for handling the time series data [35]. The time series data are considered by combining from components, i.e., a deterministic component and a stochastic component. Both components could be fitted by different models, and then models could be applied to generate new time series. However, since the sensor data usually contain multiple channels, the decomposition-based approaches may be time-consuming to fit different channels. It may also capture the correlations among different channels insufficiently [30].

Besides the statistics-based data augmentation approaches, machine learning-based data augmentation approaches empowered by generative models recently become more popular for the sensor data augmentation. For instance, Cao *et al.* proposed a statistical model as the mixture of Gaussian trees to model the time series data of minority samples [36]. In this way, the correlations between neighboring points could be learnt. However, such statistical generative models rely much on a good initial value, which is hard to be determined in real-world applications. Apart from the statistical generative models, as deep learning techniques developed rapidly, deep generative models are also proposed in recent years [37], such as generative adversarial networks (GAN) [10]. In GAN-based framework, it involves a generator and a discriminator. The generator will generate fake samples while the discriminator will distinguish whether the input samples are actual samples or fake samples. The generator and the discriminator will compete with each other until the discriminator could not distinguish the samples. At that time, the actual data distribution is learnt by the generator and the generator could be extracted out for data augmentation to obtain high-quality generated samples. In addition, compared with the conventional data augmentation approaches, GAN-based approaches could learn the data distribution, altering the augmentation process to the sampling process, which could significantly increase the number of generated samples. Hence, many GAN-based variants are developed aiming to perform the data augmentation, such as Wasserstein GAN (WGAN) [38], Least Square GAN (LSGAN) [39], and so on. However, these above-mentioned GAN-based approaches did not consider the sequential information in the data.

Notably, the generators and discriminators in GAN-based approaches are the base models rather than specific neural network structures. Hence, many GAN variants modified the neural network structures in both generators and discriminators to utilize the sequential information. However, they may still have their own shortcomings to be applied in manufacturing systems. For instance, multivariate anomaly detection with GAN (MAD-GAN) [11] is an advanced GAN-based approach for anomaly detection, but it considers the unsupervised cases without capturing label information. Besides, Time-series Generative Adversarial Networks (TimeGAN) [40] and Graph-Attention-based Generative Adversarial Network (GAT-GAN) [41] are also designed to consider temporal relationships in the data. However, it mostly considers the temporal relationship regarding features rather than a sequence, which does not fit the scenario of collecting sensor data in manufacturing systems. Hence, the generated samples quality of current GAN-based approaches may still vary by cases in manufacturing systems.

To improve the generated data quality by considering the sequential information, transformer is also commonly applied in many GAN-based frameworks for signal data augmentation [18, 19]. Compared with recurrent neural networks [42] which learn the sequence token by token, transformer learnt the entire sequence with the help of multi-head attention mechanism. The multi-head attention mechanism is essentially one kind of neural network structure, which consists of multiple heads. Each head is independent and aggregates weights on the sequence to focus on different aspects of pairwise relations among the elements. Then the information from all the heads is summarized together by matrix calculation as the overall attention score to capture the complex sequential relationships [18]. The existing works mainly consider to integrate the entire transformer into GAN-based framework, which means both generator and discriminator consists of transformers. Such architecture is complicated and still requires a large amount of training data [19]. However, since the machine is usually under normal state, the vibration signals collected under abnormal state may be insufficient for training a desired transformer model in manufacturing systems.

Overall, there are two major gaps in this study: (1) how to improve the quality of augmented sensor signals by considering the sequential information; (2) how to train the model effectively under limited data. In this study, instead of directly applying the entire transformer structure, multi-head attention mechanism is extracted and effectively integrated. In another word, the main part of the neural network structure in the generator follows the multi-head attention mechanism. In this way, the proposed model is greatly simplified but still able to consider the sequential information in the data, further improving the generated data quality. In addition, the simplified model could reduce the needed amount of training data so that it could have a good performance under limited sample size. Hence, a new data augmentation approach termed attention-stacked GAN (AS-GAN) is proposed in this study. The key novelty of the proposed AS-GAN is to propose a new attention-stacked generator, which not only integrates the multi-head attention mechanism but also stacks the multilayer perceptron. Then it could capture sequential information to improve the quality of the generated sensor signals, leading to better sensor data augmentation in manufacturing systems.

## 3. Research Methodology

This section will introduce the proposed AS-GAN in detail. In Sec. 3.1, the multi-head attention mechanism to be incorporated in GAN is presented. Afterwards, the designed attention-stacked generator for the improvement of GAN in sensor data augmentation is discussed in Sec. 3.2. Then the application of AS-GAN in advancing the classification-based manufacturing system monitoring is discussed in Sec. 3.3.

**3.1 Multi-head attention mechanism**

The multi-head attention mechanism serves as an important component in transformer [14]. It considers the pairwise relations among the elements in the input data. Hence, the complex sequential relationships could be learnt by the multi-head attention mechanism. Specifically, the multi-head attention mechanism could output the overall attention score, which is the weight to show the dependencies among different points in the sequence. Hence, it could further demonstrate the underlying relationships between each sample in a sequence, and then such relationships could be considered in data generation. Thus, it motivates this study enabling GAN to equip the strength of multi-head attention mechanism.

To formulate this framework, denote the input data as $\mathbf{X}_{n \times v}$, which have $n$ vectors with $v$ variables. The multi-head attention mechanism consists of multiple single heads. Suppose that there are $h_e$ single heads in the multi-head attention mechanism. $h_e$ is a tuning parameter which could be determined by trial experiments. As shown in Figure 2 (a), all the single heads have similar structures with the help of three weight matrices, i.e., $\mathbf{W}_i^Q$, $\mathbf{W}_i^K$ and $\mathbf{W}_i^V$.

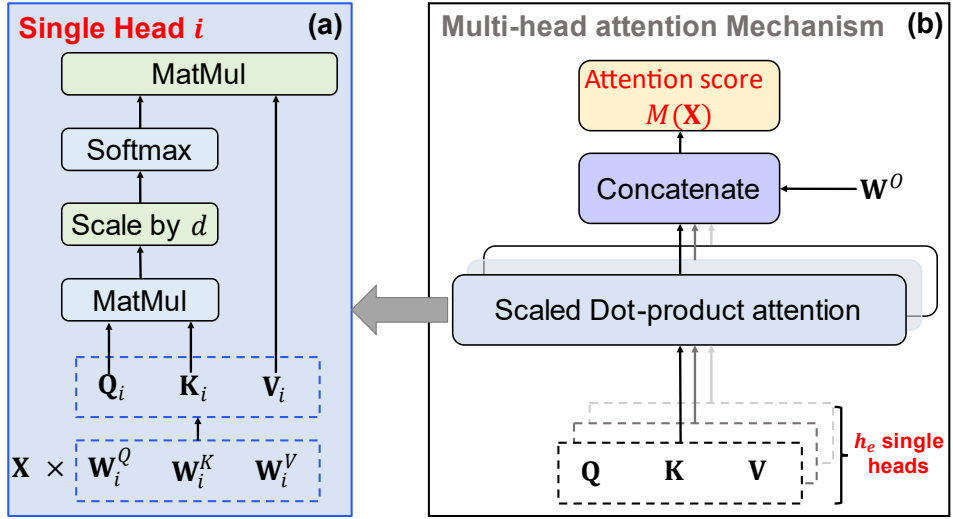

*Figure 2: The framework of single head (a) and the multi-head attention mechanism (b).*

$\mathbf{W}_i^Q$, $\mathbf{W}_i^K$ and $\mathbf{W}_i^V$ are randomly initialized and used to calculate the mapping elements of queries, keys and values, i.e., $\mathbf{Q}$, $\mathbf{K}$ and $\mathbf{V}$, respectively. Then for each single-head attention mechanism, it has an own individual direction to learn the sequential information in the data. That is, $\mathbf{A}_i$ could be obtained as the attention score of head $i$ as shown in Eq. (1), where $d = n \times v$. Afterwards, $h_e$ single heads are integrated together by concatenating the outputs from different heads, i.e., $\{\mathbf{A}_1, \mathbf{A}_2, \mathbf{A}_3, \ldots, \mathbf{A}_h\}$. Then the overall attention score, i.e., MultiHead($\mathbf{Q}, \mathbf{K}, \mathbf{V}$), could be obtained by multiplying another randomly initialized weight matrix $\mathbf{W}^O$, as shown in Eq. (1). In this way, the output of the multi-head attention mechanism, i.e., $M(\mathbf{X})$, could be demonstrated in Figure 2 (b).

$$\mathbf{Q}_i = \mathbf{X}\mathbf{W}_i^Q \;;\; K_i = \mathbf{X}\mathbf{W}_i^K ; \mathbf{V}_i = \mathbf{X}\mathbf{W}_i^V$$

$$\mathbf{A}_i = \text{softmax}\left(\frac{\mathbf{Q}_i \mathbf{K}_i^T}{\sqrt{d}}\right)\mathbf{V}_i \qquad (1)$$

$$M(\mathbf{X}) = M(\mathbf{Q}, \mathbf{K}, \mathbf{V}) = \text{MultiHead}(\mathbf{Q}, \mathbf{K}, \mathbf{V}) = \text{Concat}(\mathbf{A}_1, \mathbf{A}_2, \ldots, \mathbf{A}_{h_e})\mathbf{W}^O$$

The overall algorithm for multi-head attention mechanism is shown in Algorithm 1. Initially, the weight matrices are randomized so that different heads may capture different information among the samples in each sequence. Then the weight matrices assist to calculate the attention score in single heads and output attention scores as $\mathbf{A}_i$. Afterwards, as described in Eq. (1), the overall attention score is obtained by concatenating all the single heads as MultiHead($\mathbf{Q}, \mathbf{K}, \mathbf{V}$), i.e., $M(\mathbf{X})$. Then based on the overall attention score, the underlying relationship between different samples in the sequence, i.e., the sequential information, could be captured. Notably, the overall attention scores show the dependencies among different points in the sequence, but they are not the generated samples. Hence, another neural structure which could be applied to generate samples, i.e., multilayer perceptron (MLP), is integrated in the proposed attention-stacked generative adversarial network (AS-GAN), which is described in Sec. 3.2.

---

**Algorithm 1:** Multi-head Attention Mechanism

---

**Input:** Actual data $\mathbf{X}$, Parameter $d$ and $h_e$
**For** $i = 1$ to $h_e$ **do**
  **Step 1:** Randomly initialize or update the weight matrices $\mathbf{W}_i^Q$, $\mathbf{W}_i^K$ and $\mathbf{W}_i^V$
  **Step 2:** Calculate $\mathbf{A}_i$ by $\mathbf{W}_i^Q$, $\mathbf{W}_i^K$, $\mathbf{W}_i^V$ and $\mathbf{X}$
  **Step 3:** Randomly initialize or update the weight matrix $\mathbf{W}^O$
  **Step 4:** Concatenate $\mathbf{A}_1, \mathbf{A}_2, \ldots, \mathbf{A}_{h_e}$ together and calculate $M(\mathbf{X})$ by $\mathbf{W}^O$
**Output** $M(\mathbf{X})$

---

## 3.2 Attention-stacked GAN (AS-GAN)

As described in Sec. 2, the key idea of generative adversarial network (GAN) is to train two neural networks, namely, a generator $G$ and a discriminator $D$. $G$ will generate artificial samples $G(\mathbf{Z})$ by the input noise $\mathbf{Z}$ while $D$ will distinguish whether the input is from the actual samples $\mathbf{X}$ or artificial samples generated by $G(\mathbf{Z})$. $G$ and $D$ will compete with each other with a minimax game for $V(D,G)$, as demonstrated in Eq. (2). When $D$ is not able to classify whether the input samples are actual samples or artificial samples, $G(\mathbf{Z})$ should be much similar to the actual samples $\mathbf{X}$. At that time, the distribution of generated data, $P_{G(\mathbf{Z})}$, should be close to the underlying distribution of actual data, $P_{\text{data}}$. In this study, notably, Wasserstein GAN (WGAN) is selected as the base GAN framework for the proposed method due to its capability to improve the training stability. By integrating the multi-head attention mechanism, the attention-stacked generator, i.e., $G_M$, is proposed in Definition 1.

$$\min_G \max_D V(D,G) = \mathbb{E}_{\mathbf{X} \sim P_{\text{data}}(\mathbf{X})}[\log(D(\mathbf{X}))] + \mathbb{E}_{\mathbf{Z} \sim P(\mathbf{Z})}[\log(1 - D(G(\mathbf{Z})))] \qquad (2)$$

**Definition 1.** *(Attention-stacked Generator)*: The attention-stacked generator $G_M$ consists of a multi-head attention mechanism $M$ and an MLP-based regular generator $G$. Initially, the actual data $\mathbf{X}$ are sent to $M$ to get the overall attention score $M(\mathbf{X})$. Afterwards, both $M(\mathbf{X})$ and the noise $\mathbf{Z}$ serve as the input to $G$, and then the artificial data $G_M(\mathbf{Z}, \mathbf{X})$ can be generated. Thus, $G_M$ can be represented as

$$G_M(\mathbf{Z}, \mathbf{X}) = G(\mathbf{Z}, M(\mathbf{X})) \qquad (3)$$

More specifically, Figure 3 further demonstrates how to feed the attention score $M(\mathbf{X})$ and the noise $\mathbf{Z}$ together into the MLP-based $G$. First of all, in the first layer of MLP, the noise $\mathbf{Z}$ and the attention score $M(\mathbf{X})$, are concatenated together as $\widehat{\mathbf{X}}$ for following layers, as shown in Figure 3 (b). Subsequently, in this first fully-connected layer, also as illustrated in Figure 3 (b), the combined input $\widehat{\mathbf{X}}$ will be updated by $\widehat{\mathbf{X}} = \text{ReLu}(\mathbf{W}_1 \widehat{\mathbf{X}} + \boldsymbol{b}_1)$, where $\mathbf{W}_1$ is the randomly-initialized weight matrix, $\boldsymbol{b}_1$ is the randomly-initialized bias matrix and ReLu is the activation function for the first fully-connected layer. Then the updated $\widehat{\mathbf{X}}$ will be sent into the following fully-connected layers for calculation.

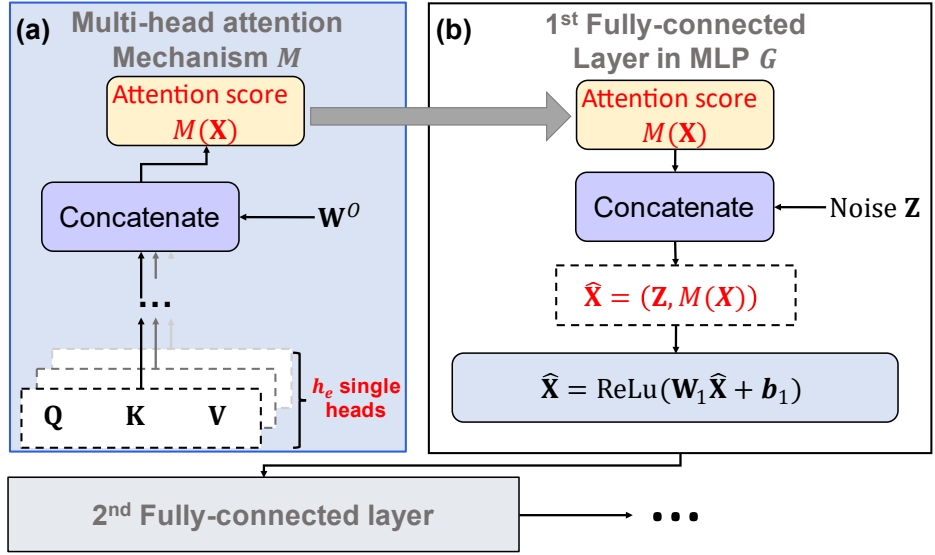

Figure 3: A demonstration showing how the attention score (a) feeding to MLP (b).

Denote that the MLP in $G$ consists of $h_f$ fully-connected layers. Similar to $h_e$ (the number of single heads,

introduced in Sec. 3.1), $h_f$ is also a tuning hyper-parameter, which could be determined by trial experiments or other hyper-parameter tuning techniques (e.g., Bayesian optimization). Notably, in the following fully-connected layer $i$ before the last layer (i.e., $1 < i < h_f$), the input $\widehat{\mathbf{X}}$ from the layer $i-1$ will be updated by $\widehat{\mathbf{X}} = \text{ReLu}(\mathbf{W}_i\widehat{\mathbf{X}} + \boldsymbol{b}_i)$, where $\widehat{\mathbf{X}}$ is obtained from layer $i-1$. Similar to the first fully-connected layer, $\mathbf{W}_i$ is the randomly-initialized weight matrix, $\boldsymbol{b}_i$ is the randomly-initialized bias matrix and ReLu is the activation function for layer $i$. However, in the last layer, i.e., the layer $h_f$, the activation function will be changed to sigmoid and the output will be the generated samples $G_M(\mathbf{Z}, \mathbf{X})$.

By integrating $G_M$ to GAN, the overall framework of the proposed AS-GAN for augmenting online sensor data in manufacturing systems is illustrated in Figure 4. With multi-head attention, different single heads are stacked together to capture the sequential information in the sensor signals and obtain the overall attention score. Then the MLP is employed since multiple fully-connected layers are stacked together. As shown in Definition 1 and Figure 3, both the noise and the overall attention score are sent to MLP for generations. Hence, the multi-head attention mechanism $M$ and the MLP $G$ can be effectively integrated in the proposed $G_M$.

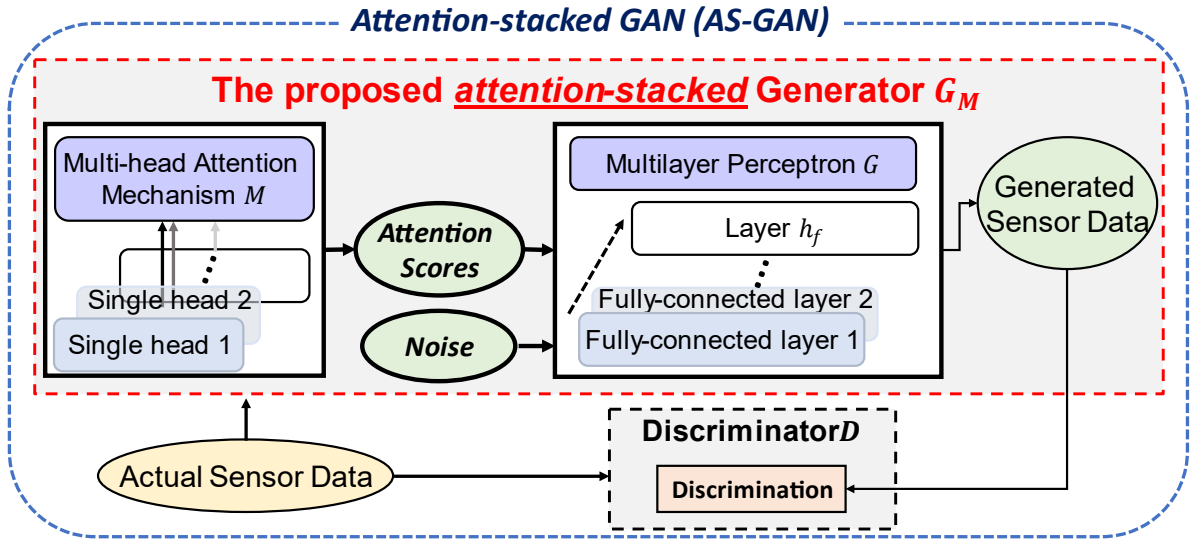

*Figure 4: An illustration of the proposed AS-GAN.*

Under such circumstances, the loss of $G_M$ and $D$, namely, $L_{G_M}$ and $L_D$, respectively, can be reformulated as

$$L_{G_M} = -\mathbb{E}_{\mathbf{Z} \sim P(\mathbf{Z})}(D(G_M(\mathbf{Z}, \mathbf{X})))$$

$$L_D = \mathbb{E}_{\mathbf{Z} \sim P(\mathbf{Z})}(D(G_M(\mathbf{Z}, \mathbf{X}))) - \mathbb{E}_{\mathbf{X} \sim P_{\text{data}}}(D(\mathbf{X})) \quad (4)$$

Due to the employed $G_M$, it is also essential to prove that, as shown in Proposition 1, the convergence property of AS-GAN is still similar as GAN.

***Proposition 1.*** Denote that actual data is from distribution $P_{\text{data}}$ and artificial data in $G_M$ is from distribution $P_{G_M(\mathbf{Z})}$. Then the model will converge, i.e., the losses will be minimized, by achieving $P_{\text{data}} = P_{G_M}$ when the minimax game in AS-GAN is formulated as

$$\min_{G_M} \max_D V(D, G_M) = \mathbb{E}_{\mathbf{X} \sim P_{\text{data}}(\mathbf{X})}[\log(D(\mathbf{X}))] + \mathbb{E}_{\mathbf{Z} \sim P_{G_M}(\mathbf{Z})}[\log(1 - D(G_M(\mathbf{Z}, \mathbf{X})))] \quad (5)$$

***Proof.*** When $G_M$ is fixed, the optimal discriminator $D$ is shown as [10]

$$D^*_{G_M}(\mathbf{X}) = \frac{P_{\text{data}}(\mathbf{X})}{P_{\text{data}}(\mathbf{X}) + P_{G_M}(\mathbf{X})} \quad (6)$$

Then the minimax game could be reformulated as

$$\max_D V(D, G_M) = \mathbb{E}_{\mathbf{X} \sim P_{\text{data}}(\mathbf{X})}\left[\frac{P_{\text{data}}(\mathbf{X})}{P_{\text{data}}(\mathbf{X}) + P_{G_M}(\mathbf{X})}\right] + \mathbb{E}_{\mathbf{X} \sim P_{G_M}(\mathbf{Z},\mathbf{X})}\left[\frac{P_g(\mathbf{X})}{P_{\text{data}}(\mathbf{X}) + P_{G_M}(\mathbf{X})}\right] \quad (7)$$

The global minimum value of $\max_D V(D, G_M)$ shown in Eq. (7) can be achieved if and only if $P_{\text{data}} = P_{G_M}$ [10]. Under this circumstance, $P_{G_M}$ can gradually converge to $P_{\text{data}}$ as the generator and the discriminator update consecutively. Hence, when the model converges, a naïve GAN could be enough for data augmentation theoretically since the generated data follows $P_{\text{data}}$ at that time. However, due to the limited sample size of the sensor data, it is usually hard to satisfy the model convergence perfectly [6]. Hence, the attention-stacked generator $G_M$ could help to make $P_{G_M}$ much closer to $P_{\text{data}}$ in practice.

The overall algorithm for the proposed AS-GAN is shown in Algorithm 2. The actual sensor data $\mathbf{X}$ are initially sent to AS-GAN. Afterwards, the overall attention score is calculated in $M$ for each batch and combined with noise $\mathbf{Z}$ to $G$. Then the output of $G_M$, i.e., $G_M(\mathbf{Z}, \mathbf{X})$, are sent to the discriminator $D$. Based on the calculated losses, both the randomly initialized matrices in $M$ and $G$ are updated. After the model is well-trained, $G_M$ could be extracted out for data augmentation. Then the augmented data could be applied to improve the online monitoring performance of manufacturing systems, which is illustrated in Sec. 3.3.

---
**Algorithm 2:** AS-GAN
---
**Input:** Actual data $\mathbf{X}$, Parameter $d$, $h_e$, $h_f$, iteration times $t$, batch size $s$
**For** $j = 1$ **to** $t$ **do**
   **Step 1:** Randomly choose $s$ samples as $\mathbf{X}_j$ from $\mathbf{X}$
   **Step 2:** Obtain overall attention score $M(\mathbf{X}_j)$ by sending $\mathbf{X}_j$, $d$ and $h_e$ to $M$
   **Step 3:** Randomly initialize or update $\mathbf{W}_1$ and $b_1$
   **Step 4:** Concatenate $M(\mathbf{X}_j)$ and randomly generated noise $\mathbf{Z}$ as $\widehat{\mathbf{X}}$
   **Step 5:** Update $\widehat{\mathbf{X}}$ by $\mathbf{W}_1$ and $b_1$
   **For** $k = 2$ **to** $h_e$ **do**
     **Step 6:** Randomly initialize or update $\mathbf{W}_k$ and $b_k$
     **Step 7:** Calculate and update $\widehat{\mathbf{X}}$ by $\mathbf{W}_k$ and $b_k$
**Step 8:** Output $\widehat{\mathbf{X}}$ as $G_M(\mathbf{Z}, \mathbf{X}_j)$
**Step 9:** Send $G_M(\mathbf{Z}, \mathbf{X}_j)$ and $\mathbf{X}_j$ into discriminator $D$ to get output $D(G_M(\mathbf{Z}, \mathbf{X}_j))$, $D(\mathbf{X}_j)$
**Step 10:** Update $G_m$ and $D$ by the calculated losses $L_{G_M}$ and $L_D$, respectively
**Until** $L_{G_M}$ and $L_D$ converge:
   Extract $G_M$ out for data augmentation

---

### 3.3 AS-GAN for sequential data augmentation in imbalanced classification

As described in Sec. 2.1, the insufficient abnormal sensor signals may lead to significant bias in monitoring. Then the proposed AS-GAN synthesizes the abnormal sensor signals to improve the monitoring performance of manufacturing systems. Hence, how to incorporate the augmented abnormal sensor signals in the monitoring process is identified in this section. According to recent literatures [23-25], the supervised machine learning approaches, particularly, the classification-based approaches, are extensively utilized in online quality monitoring, where there is a strong demand for effective data augmentation approaches. Based on Proposition 1 in Sec. 3.2, the distribution of augmented sensor signals should be the same as the distribution of actual sensor signals. Hence, the performance classification approaches can be potential much better since the augmented sensor signals could significantly improve the classification performance.

The overall framework to apply AS-GAN in classification is shown in Figure 5. Since the collected sensor signals are sequential data, both normal and abnormal sensor signals are organized in a window-based format. Afterwards, the window-based abnormal sensor signals are sent to AS-GAN for training. Then the sequential information in each window-based abnormal sensor signal could be captured and learnt in the AS-GAN. When the AS-GAN model is well-trained, the artificial sensor signals are generated as augmented signals under abnormal state. Then the artificial abnormal sensor signals are combined with the imbalanced actual sensor signals to make the entire dataset balanced. Such balanced dataset is considered as a training set to train the classifier.

After the classifier is well-trained, new actual sensor signals can be sent to the classifier and detect whether anomalies exist or not. It is important to note that, when the proposed AS-GAN model is well-trained, the attention-stacked generator could generate a large number of samples simultaneously. Hence, the computation efficiency of the proposed method is sufficient to monitor manufacturing systems in real time whether anomalies appear. Besides, with the help of the augmented abnormal sensor signals, the classifier should be more accurate than the classifier trained by imbalanced actual sensor signals, which could further improve the online monitoring performance to determine the state of the manufacturing systems.

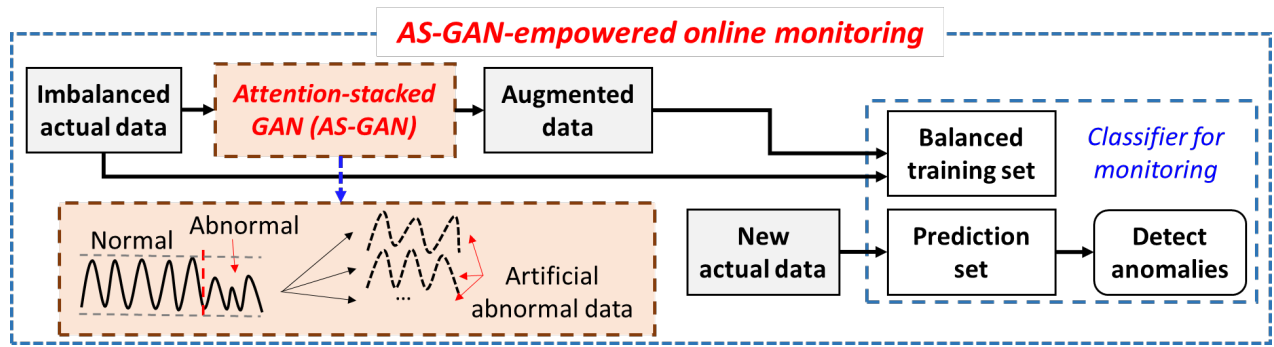

*Figure 5: An illustration of online monitoring application in manufacturing empowered by the proposed AS-GAN.*

In addition, such online monitoring application could also help to demonstrate the quality of augmented sensor signals, and further demonstrate the effectiveness of the proposed AS-GAN. If the proposed method is effective, the augmented sensor signals should be similar to the actual sensor signals under the abnormal conditions. Afterwards, the classification metrics comparing with the metrics collected from the classifier which is trained without data augmentation should be better, e.g., the classification accuracy is expected to be higher. Then it could prove that the proposed method could synthesize high-quality sensor signals to improve the monitoring performance.

## 4. Real-world Case Study in Additive Manufacturing

### 4.1 Experimental setup

To validate the effectiveness of the proposed AS-GAN, a real-world case study in additive manufacturing (AM) process is designed and applied in this work (Figure 6). According to the design mechanism of AM, the printing is in the layer-by-layer manner [43]. However, due to the defects or unintended anomalies which may occur during the printing process [44], the labor and financial cost may increase. Such defects or unintended anomalies could be reflected by the collected sensor signals [45], which could be applied to perform online monitoring. However, due to the insufficient data collected under abnormal state, data augmentation is needed.

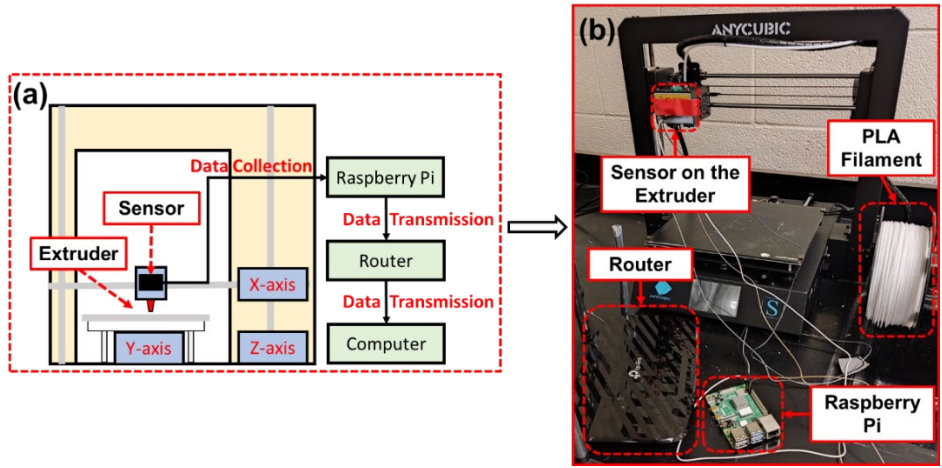

*Figure 6: The designed validation platform (a) and the physical experimental platform setup (b).*

In this case, as shown in Figure 6(a), the experimental data was designed to be collected from a regular fused filament fabrication (FFF) machine. Specifically, there is one accelerometer sensor attached on the extruder to collect the sensor signals of the X-axis extruder acceleration. To simulate the scenarios of limited sample size, the sampling frequency was controlled as approximately 1 Hz. The Raspberry Pi 4b microcontroller was used for data acquisition from the accelerometer. Then the collected data are sent to the computer for analysis through the router which enables a cyber-enabled environment. Following the designed validation platform, the physical experimental platform setup is demonstrated in Figure 6 (b). During each printing, there are about 850 observations of one channel as the sensor signals collected by time.

Through polylactic acid (PLA) filament, a solid cube was printed with dimension 2×2×2 cm$^3$ as shown in Figure 7 (a). In addition, a small square void was maliciously inserted on the cube within the red solid line as shown in Figure 7 (b). Hence, the vibration signals are under normal state when the layers are printed without the square void, and are under abnormal state when the layers are printed with the square void.

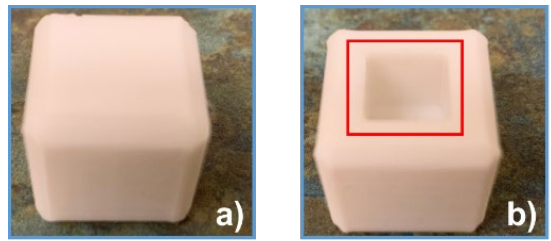

Figure 7: Sample cubes with normal (a) and anomaly (part within red solid line) (b).

Five different trials of this case have been conducted. The collected sequential data from trial 1 and trial 2 is demonstrated as an example in Figure 8, where x-axis represents the timestamps and y-axis represents the acceleration of the extruder. Approximately, the first 5%~35% collected data are extracted as normal samples in this case while the latter 65%~75% collected data are extracted as abnormal samples. Among the timestamps within red dashed line, the printer is printing the cube as shown in Figure 7 (a), where the collected signals are considered under normal state. Among the timestamp within green dashed line, the printer is printing the cube as shown in Figure 7 (b), where the collected signals are considered under abnormal state. Specifically, there is no significant differences between the scale and appearance of signals collected under normal and abnormal state. Hence, to identify whether the machine is under normal or abnormal state, a classifier needs to be applied to distinguish the collected data.

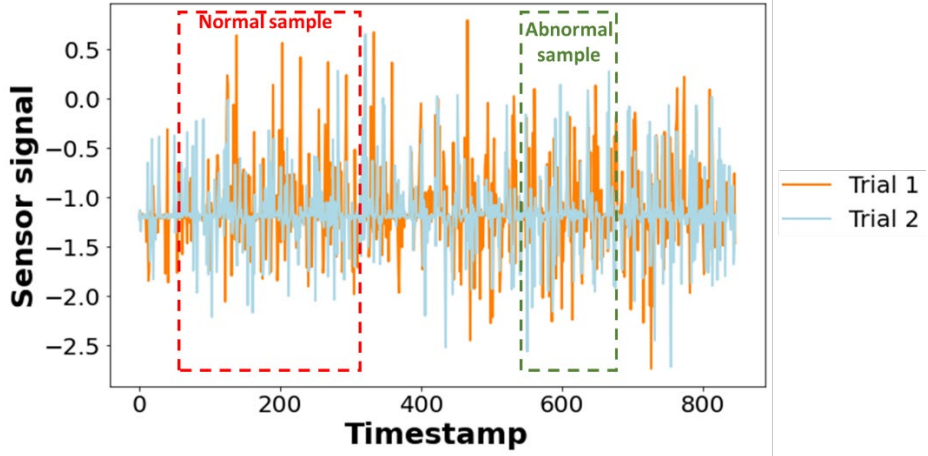

*Figure 8: The collected sensor signals from trial 1 and trial 2.*

To capture the sequential relationships in the data, window-based sampling is applied. Recall that the input data is denoted as $\mathbf{X}_{n \times v}$ in Sec. 3.1. In this case, $n$ is set as 30 and $v$ is 1. That is, the window size is 30 and the collected data has one channel. In order to increase the number of window-based samples, the overlap size between adjacent windows is 28. Then there are about 40 abnormal window-based samples and 120 normal window-based samples from the collected sensor signals.

In addition, as shown in Figure 8, the collected sensor signals have no significant differences between different trials. Therefore, to fully validate the effectiveness of the proposed method, different trials are applied separately as training or testing set. Trial 1 is applied as the training set to train the proposed method and the classifier, while trial 2 to trial 5 are applied as the testing set for the classifier. As shown in Figure 6, the normal and abnormal samples are too close so that the basic machine learning classifier, such as random forest [46] and support vector machine [47], are not satisfying according to preliminary trials. Hence, the convolutional neural network (CNN) classifier [48] is applied as the classifier for online monitoring to have the reasonable baseline performance, which contains two convolutional layers and three fully-connected layers. As discussed in Sec. 3.3, if the proposed data augmentation approach is effective, the quality of the generated artificial sensor signals should be high. Afterwards, with the same testing set, the classifier trained by both the artificial sensor signals and actual sensor signals should have the better performance than the classifier trained by actual sensor signals.

## 4.2 Parameter selection and benchmark methods

In this study, it is important to initially talk about the parameter selection. There are two parameters to be determined, i.e., $h_e$ and $h_f$. To determine $\{h_e, h_f\}$, each pair of $\{h_e, h_f\}$ from $h_e = 1, 3, 5, 10$ and $h_f = 1, 3, 5, 10$ are selected. To select the optimal pair of $\{h_e, h_f\}$, one classification metric needs to be calculated. Thus, the F-score [49], which could show the overall classification performance, is applied to demonstrate the performance of the proposed AS-GAN under different pairs of $\{h_e, h_f\}$. The pair of $\{h_e, h_f\}$ with the highest F-score of the proposed method will be chosen. The baseline F-score for trial 2 to trial 5 are 0.606, 0.797, 0.717, 0.736, respectively. Since the baseline F-score of trial 3 to trial 5 are much similar and more convincing, the proposed method will tune the parameter under trial 3 to trial 5. Then the average F-scores will be calculated for parameter selection. The discriminator of AS-GAN is initially designed as a two-layer MLP and the number of iterations is set as 7,000. In each trial, 80 artificial abnormal samples will be generated and sent to the CNN classifier as part of the training set. Then the training set for the classifier may have the same number of normal and abnormal sensor signals.

The average F-scores and its standard deviations for each pair of $\{h_e, h_f\}$ are shown in Table 1. Under each specific $h_e$, the F-scores will significantly decrease when $h_f$ increases, which shows that one fully-connected layer is already able to generate the sensor signals accurately. Especially, when $\{h_e, h_f\} = \{3, 1\}$, it is clearly shown that the proposed method has the highest F-score. Though the F-score under $\{h_e, h_f\} = \{5,1\}$ are also much similar to the highest one, the standard deviation of $\{h_e, h_f\} = \{3, 1\}$ is much lower. Therefore, the pair of $h_e = 3$ and $h_f = 1$ is selected in this study. That is, the attention-stacked generator will contain three heads for multi-head attention mechanism and one fully-connected layer for MLP.

***Table 1:*** *The F-scores and standard deviations under different pairs of $\{h_e, h_f\}$*

| $h_e$ | $h_f$ | | | |
|---|---|---|---|---|
| | 1 | 3 | 5 | 10 |
| 1 | 0.795 (0.022) | 0.790 (0.079) | 0.679 (0.079) | 0.650 (0.036) |
| 3 | **0.875 (0.023)** | 0.800 (0.059) | 0.794 (0.072) | 0.651 (0.081) |
| 5 | 0.874 (0.043) | 0.732 (0.083) | 0.653 (0.036) | 0.655 (0.045) |
| 10 | 0.819 (0.051) | 0.840 (0.055) | 0.810 (0.057) | 0.680 (0.139) |

To validate the effectiveness of the proposed AS-GAIN comprehensively, benchmark methods are essential for comparison. Since the multi-head attention mechanism is mainly applied in transformer, it is important to demonstrate why only the multi-head attention mechanism is applied in the proposed method rather than the transformer framework. Hence, transformer-based GAN (T-GAN) is applied as one of the benchmark methods. Particularly, T-GAN incorporates the transformer framework in the generator while the other setups are the same as AS-GAN. Besides, SMOTE, GAN, and WGAN are also selected as benchmarks.

To ensure the fairness of comparison, the above-mentioned benchmark methods will have the similar parameter setup as AS-GAN. As described in Sec. 4.1, trial 1 will be applied as the training set for both AS-GAN and the classifier, while trial 2 to trial 5 will be applied as the testing set for the classifier. In addition, to make the results more representative, each experiment involves three replicates and then the average F-scores (with standard deviations) are used for comparison.

**4.3 Result discussion**

Following the setup in Sec. 4.2, F-score comparisons between the proposed method and benchmark approaches are shown in Table 2. For each benchmark approach, it may occur that the F-score after data augmentation is smaller than the baseline. However, the proposed method could always improve the classification performance under each trial. Hence, it shows that the proposed method could generate artificial samples accurately, i.e., the quality of generated sensor signals is stably better than other benchmark approaches.

*Table 2: The F-scores and standard deviations for the proposed method and benchmark methods*

| Methods | Trials | | | |
|---|---|---|---|---|
| | Trial 2 | Trial 3 | Trial 4 | Trial 5 |
| Baseline | 0.606 (0.013) | 0.797 (0.013) | 0.717 (0.015) | 0.736 (0.011) |
| SMOTE | 0.612 (0.024) | 0.723 (0.020) | 0.761 (0.019) | 0.706 (0.021) |
| GAN | 0.612 (0.003) | 0.770 (0.002) | 0.741 (0.002) | 0.654 (0.003) |
| WGAN | 0.607 (0.003) | 0.678 (0.002) | 0.678 (0.002) | 0.625 (0.003) |
| T-GAN | 0.519 (0.055) | 0.778 (0.004) | 0.756 (0.004) | 0.899 (0.008) |
| **AS-GAN (Proposed)** | **0.650 (0.013)** | **0.873 (0.009)** | **0.854 (0.005)** | **0.899 (0.005)** |

Comparing SMOTE with the proposed AS-GAN, the F-scores of AS-GAN are significantly higher than the F-scores of SMOTE. Specifically, the standard deviations of SMOTE are also much higher. Therefore, it demonstrates that the sequential information in the collected sensor signals is vital for data augmentation rather than the linear relationship captured by SMOTE. In addition, the F-scores of GAN and WGAN are always smaller than the F-scores of AS-GAN under each trial, which also validates the effectiveness of attention-stacked generator.

Besides, the F-scores of the proposed method are also significantly higher than the F-scores of T-GAN under trial 2 to trial 4. Though T-GAN could achieve the same F-score under trial 5, the standard deviation of T-GAN is higher than the standard deviation of the proposed AS-GAN. Thus, the multi-head attention mechanism is more suitable for the attention-stacked generator rather than transformer in this paper. Overall, the F-scores of the proposed method are always the highest under each trial, which shows the effectiveness of the proposed method for window-based data augmentation. It is important to note that, each weight matrices in multi-head attention mechanism is randomly initialized for each replicate. Hence, according to the relatively small standard deviations of the proposed method, the randomization of weight matrices will not influence the performance of the proposed method significantly.

In addition, if the proposed method is effective, all the generated samples should be more similar to the actual abnormal samples rather than the actual normal samples. Hence, the average Euclidean distance between each augmented abnormal sample $X_{Augmented}$ and all actual abnormal samples $\mathbf{X}_{Abnormal}$, denoted by $d_1 = d(\mathbf{X}_{Abnormal}, X_{Augmented})$, as well as the average Euclidean distance between each augmented abnormal sample $X_{Augmented}$ and all actual normal samples $\mathbf{X}_{Normal}$, denoted by $d_2 = d(\mathbf{X}_{Normal}, X_{Augmented})$, are calculated for comparison, respectively.

As shown in Figure 9, 100 samples generated for the abnormal states are used to perform this validation experiment. In the figure, each blue triangle dot represents a point $(d_2, d_1)$ for each generated sample. To better visualize the results, an orange dashed slash line which shows the line $d(\mathbf{X}_{Abnormal}, X_{Augmented}) =$ $d(\mathbf{X}_{\text{Normal}}, \mathbf{X}_{\text{Augmented}})$ was also plotted (i.e., line $d_1 = d_2$). The results clearly show that most of the points are beneath the orange line, meaning that $d(\mathbf{X}_{\text{Abnormal}}, \mathbf{X}_{\text{Augmented}}) < d(\mathbf{X}_{\text{Normal}}, \mathbf{X}_{\text{Augmented}})$, i.e., $d_1 < d_2$, for more of the generated sample. Therefore, such pattern demonstrates that the generated samples are closer to actual abnormal samples, which also shows the effectiveness of the proposed method to generate high-quality abnormal state samples.

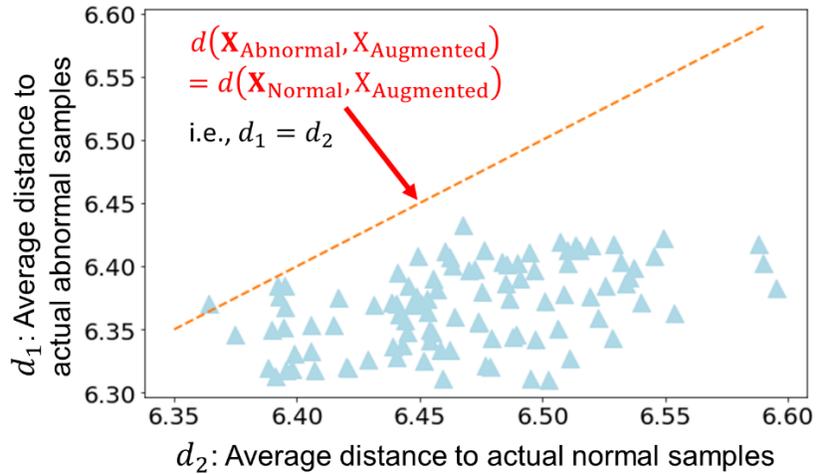

*Figure 9: The comparison between $d_1$ and $d_2$ for 100 generated abnormal state samples. For the generated abnormal state data, they are closer to the actual abnormal samples than the normal samples.*

In addition, to further demonstrate the robustness of the proposed method, the experiments are conducted under different balanced ratios, i.e., the ratio of the number of abnormal sensor signals to the number of normal sensor signals. The balanced ratio is selected from the set {0.07, 0.13, 0.20, 0.27, 0.33}. Under each balanced ratio, the number of generated samples for the proposed method and benchmark methods are the same to achieve that the balanced ratio in the training set of the classifier is 1. The other parameter setups are the same as previous setups.

The F-scores for trial 2-5 under different balanced ratios are shown in Figure 10. Figure 10 (a) demonstrates the F-scores when using trial 2 as testing set. Though T-GAN may have a better performance when the balanced ratio is 0.20, it has an extremely low F-score when the balanced ratio is 0.33, demonstrating the low robustness of the T-GAN. Compared to T-GAN, the proposed method mostly has the highest F-scores without any significant decreasing or increasing patterns. Hence, the F-scores under trial 2 could show the relatively high robustness of the proposed method.

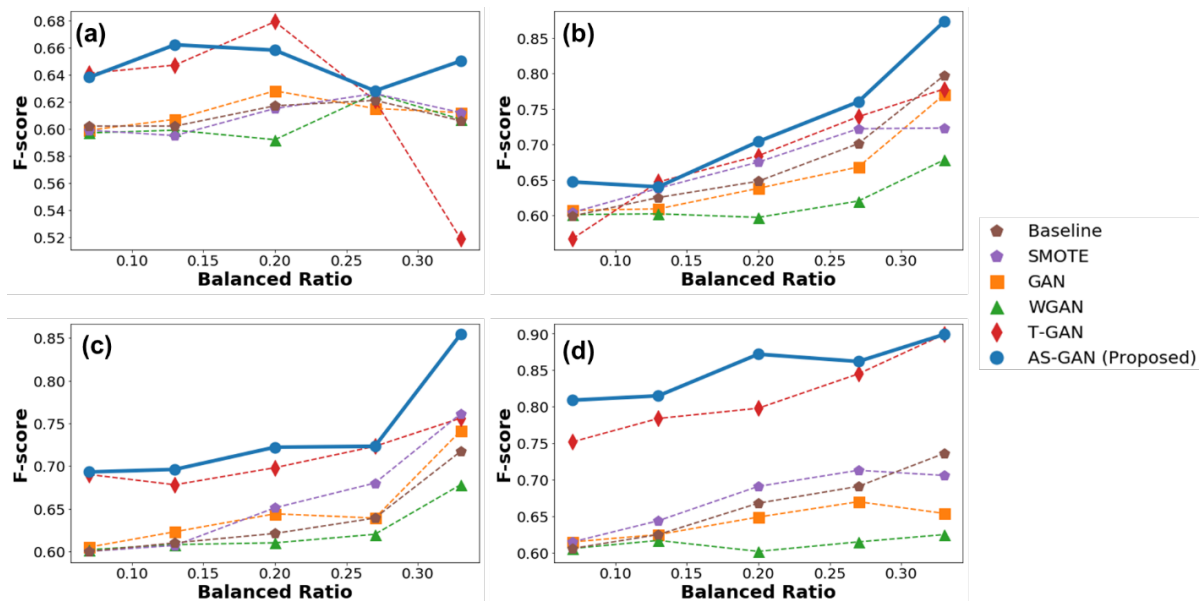

Figure 10: The F-scores for the proposed methods and benchmark approaches with balanced ratio series under different trials: (a) trial 2, (b) trial 3, (c) trial 4, and (d) trial 5.

As for Figure 10 (b)-(d), the proposed method mostly has the highest F-score under each balanced ratio and each trial. Overall, it demonstrates the outperformance of the proposed method. Besides, comparing with other benchmark approaches, the F-scores of T-GAN are also mostly higher under different balanced ratio and different trials. Hence, the outperformance of T-GAN and the proposed AS-GAN demonstrate the effectiveness of incorporating multi-head attention mechanism in the GAN-based structure. In addition, as the balanced ratio increases, the F-scores of all the approaches also gradually increase. This is reasonable since the higher balanced ratio represents more information to be learnt in data augmentation. The reason why such pattern does not occur in the F-scores calculated from trial 2 may be because the abnormal sensor signals in trial 2 are less efficient than the abnormal sensor signals from other trials. Overall, the proposed method is effective to generate accurate samples and help to improve the classification performance.

It is also important to note that, the proposed method could generate more than 20 samples in one second (i.e., more than 20Hz) in this case (by Python 3.7.4 on Intel® Core™ Processor i7-9750H (Hexa-Core, 2.60

GHz)). As described in Sec. 4.1, the sampling frequency of the AM process in this case is 1 Hz. Moreover, it is capable of handling the sensors with higher sampling frequency. Hence, the computation efficiency of the proposed AS-GAN is also sufficient enough for online monitoring.

## 5. Conclusions

In this paper, a new data augmentation approach termed AS-GAN is proposed to augment the sensor signals effectively with the consideration of sequential order in advanced manufacturing systems, which could be helpful to improve the online process monitoring performance. Compared with the existing GAN models, a multi-head attention mechanism is designed to capture the sequential information. Besides, the attention-stacked based framework is also incorporated in the generator. Furthermore, the framework to apply the proposed AS-GAN is demonstrated and discussed. Essentially, the proposed method could be incorporated into most of the common GAN models. In this study, WGAN is selected as the base model for the proposed method. The superior performance of AS-GAN over the benchmark methods is demonstrated by a real-world case study in AM. Specifically, compared with the baseline, the F-score improvements of the proposed method achieve 0.163 at most. In addition, the effectiveness of the multi-head attention mechanism and attention-stacked framework is also validated in the case study through comprehensive comparisons between the proposed method and benchmark methods. The F-scores of the proposed method are always higher than the F-scores of the benchmark methods, where the highest may achieve 0.255. Therefore, this method is promising for sensor-based data augmentation and sensor-based online monitoring in advanced manufacturing systems.